\documentclass[]{spie}  

\usepackage{amsmath,amsfonts,amssymb}
\usepackage{graphicx}
\usepackage[colorlinks=true, allcolors=blue]{hyperref}
\graphicspath{ {./images/} }

\title{Unsupervised Learning for Target Tracking and Background Subtraction in Satellite Imagery}

\author[a]{Jonathan S. Kent}
\author[a]{Charles C. Wamsley}
\author[a]{Davin Flateau}
\author[a]{Amber Ferguson}
\affil[a]{Ball Aerospace \& Technologies Corp., 2875 Presidential Drive, Fairborn, OH, USA 45324}

\authorinfo{Send correspondence to Jonathan S. Kent\\E-mail: jskent@ball.com}

\begin{document} 
\maketitle

\begin{abstract}
This paper describes an unsupervised machine learning methodology capable of target tracking and background suppression via a novel dual-model approach. ``Jekyll" produces a video bit-mask describing an estimate of the locations of moving objects, and ``Hyde" outputs a pseudo-background frame to subtract from the original input image sequence. These models were trained with a custom-modified version of Cross Entropy Loss.\\
\\
Simulated data were used to compare the performance of Jekyll and Hyde against a more traditional supervised Machine Learning approach. The results from these comparisons show that the unsupervised methods developed are competitive in output quality with supervised techniques, without the associated cost of acquiring labeled training data.
\end{abstract}


\section{INTRODUCTION}
\label{sec:intro}  

Target tracking and background subtraction are central tasks in the field of satellite imagery analysis \cite{1986ah...bookQ....B, PhysRevD.83.103514, 7326160}, made more difficult by the existence of background motion, weather patterns, and both static and transient sensor artifacts. However, manual segmentation of available data remains costly in time, effort, and money \cite{rashtchian2010collecting}, and transfers that cost to supervised learning-based projects, which require a set of manually created targets to train on. Therefore, despite the increased theoretical complexity required, it is fruitful to develop unsupervised algorithms and model architectures capable of operating in this area, which do not require explicit targets and therefore are not burdened by the cost of creating them \cite{hastie2009unsupervised}. The solution proposed here involves developing two networks, called ``Jekyll" and ``Hyde,"\footnote[1]{In the book ``The Strange Case of Dr. Jekyll and Mr. Hyde," Dr. Jekyll and Mr. Hyde are two halves of the same person, and a recurring theme involves Dr. Jekyll using his wealth and influence to mask over the crimes committed by Mr. Hyde, while they are investigated by Mr. Utterson. Similarly, Jekyll produces a bit-mask at cost to itself that covers the errors generated by Hyde, and they are compared to a supervised model named Utterson.} \cite{stevenson1947strange} which respectively produce a target tracking bit-mask over a series of frames, and an approximation of a background image. These two outputs are compared to the input used to generate them, with a new loss function defined to express the difference between the Hyde-generated pseudo-background and every frame of the input, but masked out by the presence of an object detection by Jekyll.\\
\\
The data domain in use for these experiments consists of Earth-imagery frames that change over time, i.e. ``video" from an infrared ``camera", containing targets of interest which are spatially unresolved. These targets are observed by the sensors with variable Signal-to-Noise ratios.\\
\\
The data used for the models and experiments in this work were simulated using the Air Force Institute of Technology (AFIT) Sensor and Scene Emulation Tool, or ASSET \cite{holst_development_2017, noauthor_afit_2020}. However, the work presented in this paper focuses solely on developing an initial approach to an unsupervised target tracking step, rather than providing a complete solution for target tracking and background subtraction, robust to background motion, weather, and sensor artifacts. To facilitate this, those confounding effects were left out of the data generated using ASSET. Further work will be required to address them for real world applications.\\
\\
The actual data-set generated consists of two $500 \times 500 \times 500$\footnote[1]{500 frames, 500 pixels of width and height.} data tensors: an input data tensor containing Gaussian scaled\footnote[2]{Here, Gaussian scaling of a data-set $\mathcal{D}_u$ into $\mathcal{D}_s$ consists of $\mathcal{D}_s \leftarrow \frac{\mathcal{D}_u - \mu(\mathcal{D}_u)}{\sigma(\mathcal{D}_u)}$, which gives $\mathcal{D}_s$ a mean of $0$ and a standard deviation of $1$. Where $\mathcal{D}_u$ may have values in the hundreds of thousands, which would break the numerics of neural network processing, $\mathcal{D}_s$ will be constrained relatively close to $0$.}\cite{zill2011advanced} infrared intensity ``image" or ``video" data including backgrounds and target objects, and a synthetic target data tensor consisting of a corresponding pixel-by-pixel binary target mask \cite{iliadis2016deepbinarymask} denoting the locations of target objects. The synthetic target data is used for reporting numerical and graphical results, as well as to train a supervised model that can be used as a benchmark for the unsupervised models. These tensors were carved into samples consisting of $16$ frames of $64 \times 64$ video, which were separated into Training, Validation, and Testing  subsets to avoid over-fitting during experimentation \cite{Goodfellow-et-al-2016} with 50\%, 20\%, and 30\% of the samples respectively.

\section{RELATED WORK}
\label{sec:related}

Modern developments in unsupervised learning were kick-started by the development of the Generative Adversarial Networks (GAN) framework \cite{goodfellow_generative_2014}, which was one of the first to introduce a methodology involving multiple networks ``playing against" one another, hence \textit{adversarial} networks. Typically, GANs are employed in the task of generating new samples from a high-dimensional distribution, by feeding a generator network a random vector and having a discriminator network attempt to determine which of a collection of samples have been generated artificially. Somewhat similar in nature to the GAN is the Variational Auto-Encoder (VAE) \cite{kingma_auto-encoding_2014}, which attaches the output of an encoding network directly to the input of a decoding network, and trains both according to reconstruction loss. In the case of VAEs, the two networks share a loss function and are encouraged to work together to minimize it. The use of multi-network systems for unsupervised learning inspired the approach used in this paper.\\
\\
Previous work involving VAEs have shown that it is possible to perform image segmentation in an unsupervised setting \cite{xia_w-net:_2017}. This was achieved by minimizing both the reconstruction error of the VAE, and the normalized cut \cite{shi2000normalized} of the encoded state. Although image segmentation and target tracking possess theoretical similarities, in an unsupervised setting the tendency of segmentation algorithms is towards segmenting the background into, for example, mountains and valleys, rather than background versus target objects. \cite{wang_unsupervised_2019} introduced an extremely clever unsupervised target tracking scheme that achieved high accuracy. It produces a single target track forwards and backwards through time, and takes the differences between the two tracks as its loss, assuming the only consistent solution is to track a moving target, using regularization to rule out staying put. Unfortunately, it is unsuitable for analyzing satellite imagery, as it can neither track multiple objects simultaneously, nor render the size of detected objects.\\
\\
Both \cite{ghasemi_unsupervised_2011} and \cite{nair_unsupervised_2004} accomplished unsupervised object detection using classical methods. \cite{ghasemi_unsupervised_2011} leverages algorithmic clustering of similar arrangements of pixels together, and \cite{nair_unsupervised_2004} uses a modified form of the Winnow algorithm \cite{littlestone1988learning}. Both of these predate current Computer Vision research, and as a result do not leverage the capabilities of neural Machine Learning, instead producing much less powerful linear classification models akin to a single convolutional layer. Both algorithms are limited by their problem formulation, in that they are intended to be used with a single background, e.g. a separate model would be developed for each camera in a security system because they look at different scenes around a house, despite them all being used for the same task. This leaves them both incapable of generalization, and both assume that unusual deviation from the mean value of a pixel implies an object of interest. However, by abstracting much of the algorithmic work of \cite{ghasemi_unsupervised_2011} or \cite{nair_unsupervised_2004} to the Machine Learning model, their shared assumption that unusual deviation implies interest can be reused in designing a loss function.  

\section{Model Formulation and Loss Function}
\label{sec:modformloss}
\subsection{Formulation}

Both Jekyll and Hyde are formulated as feed-forward, three dimensional convolutional neural networks (CNN) \cite{zhang1990parallel, lecun1989backpropagation, ji20123d}, implemented with PyTorch \cite{paszke2019pytorch, paszke2017automatic}. Due to the nature of the satellite sensor, the imagery that is used represents an approximate intensity mapping, scaled according to mean and standard deviation, rather than color photography, and therefore is to be treated as single-channel rather than multi-channel imagery. Both Jekyll and Hyde accept a $K \times 1 \times N \times W \times H$ tensor as their input, where $K$ is the number of samples in the batch, $N$ is the number of frames in a sample, and $W \times H$ describes the width and height of the input image.\\
\\
Both models employ a number of skip/highway connections \cite{DBLP:journals/corr/SrivastavaGS15} in their architecture, concatenating the hidden states from the $n^{th}$ layer in the first half of the network with the inputs to the $n^{th}$-from-the-last layer in the second half of the network. This helps to combat gradient degradation in the early layers, as well as giving access to more primitive data to the later layers. The structure of these networks could be analogized to an hourglass, skip connections notwithstanding. The first half of the network increases the number of filters and decreases the width and height of the hidden state, with the second half of the network reversing the process by decreasing the number of filters and increasing the width and height. The first half consists of convolution and max-pool operations, and the latter half of transpose convolutions and max-unpool operations. ReLU activations \cite{glorot2011deep} are used between convolution-type operations. This is somewhat similar to the model architectures described in \cite{ronneberger2015u, long_fully_2015, mao2016image, quan2016fusionnet}, but simplified and using 3D convolutions.\\
\\
The only architectural difference between Jekyll and Hyde is an output function, which is executed immediately after their last layer. Jekyll possesses a final Sigmoid activation function \cite{Goodfellow-et-al-2016}, turning a hidden state into a tensor of probabilities denoting the estimated likelihood that a given pixel at a given time contains a target object. Hyde's final output is taken as the frame-wise mean over its last layer, producing a single frame to be used as an approximated background.\\
\\
For direct comparison between supervised and unsupervised methods in this domain, a competing supervised model, named ``Utterson" \cite{stevenson1947strange}, is trained against synthetic target data. Utterson possesses an architecture identical to that of Jekyll, and is tested on the same task as Jekyll; producing a bit-mask denoting the positions of targets of interest. For a loss function, Utterson uses vanilla Binary Cross Entropy loss, as is natively supported in PyTorch \cite{Goodfellow-et-al-2016, paszke2019pytorch}. A supervised model against which to compare the unsupervised models will allow for direct analysis of the costs and benefits associated with using unsupervised instead of supervised learning on this task.



\subsection{Loss Function}

The goal of the loss function is to impel Hyde to produce an approximation of the background of the input image sequence sans moving target objects, and get Jekyll to label the locations of those moving objects without also including portions of the background. This will be accomplished by using a modified form of Cross Entropy Loss \cite{Goodfellow-et-al-2016} along with a cost associated with masking. While Cross Entropy Loss is typically used for classification problems, by formulating target tracking as ``classifying pixels between is-a-target and just-the-background," it becomes applicable.\\
\\
The typical formulation of Binary Cross Entropy Loss is given as: $$-\sum_{x \in \mathcal{X}} Y(x) \textbf{ln}\big(\hat{Y}(x)\big) \ + \ \big(1 - Y(x)\big) \textbf{ln}\big(1 - \hat{Y}(x)\big)$$ where $Y(x)$ is the true label of $x$, and takes the value $1$ where $x$ is a target, and $0$ where $x$ is the background. $\hat{Y}(x)$ is the probability estimated by the model that $x$ is a target. By using the squared difference between the input and Hyde's pseudo-background as an analogue for $Y(x)$, and replacing the $\big(1 - Y(x)\big) \textbf{ln}\big(1 - \hat{Y}(x)\big)$ term\footnote[1]{This term breaks down for $Y(x) > 1$, and so requires modification. The original term's use was that it would increase loss where the model incorrectly output high values, encouraging it not to simply declare everything to be a target. But it will be demonstrated that, in this case, this functionality can be replicated by assigning a constant cost to outputting any value at all, encouraging the model only to ``spend" where it expects to lower the other term by a greater amount.} with a simpler cost term, an alternate version of entropic loss can be used in the unsupervised case. In addition, Hyde contributes to decreasing loss by producing a background that minimizes the difference between it and the input, except where Jekyll is masking it out. As a result, the same loss can be used to optimize both models. This loss function will now be constructed from its elements.\\
\\
For the purposes of simplifying notation, the loss function is described as it applies to a single $N \times W \times H$ sample, rather than to the full $K \times 1 \times N \times W \times H$ batch input.\\
\\
\noindent The tensor describing the frame-wise linear differential between Hyde's output $H$ and the input $i$, the frames of which are $i_1, i_2\ ... \ i_N$:
\begin{center}
$\hat{\Delta}(\theta_h, i) = \begin{bmatrix} i_1 - H(\theta_h, i), & i_2 - H(\theta_h, i), & \dots & i_N - H(\theta_h, i)\end{bmatrix}$
\end{center}
\begin{flushleft}
The element-wise square of the linear differential:
\end{flushleft}
\begin{center}
$\hat{\Delta}^2(\theta_h, i) = \hat{\Delta}(\theta_h, i) \odot \hat{\Delta}(\theta_h, i)$
\end{center}
\begin{flushleft}
The tensor containing the squared differentials, masked out by the negative natural logarithm of Jekyll's bit-mask:
\end{flushleft}
\begin{center}
$\hat{\mathcal{L}}_h(\theta_h, \theta_j, i) = -\textbf{ln}\big(J(\theta_j, i) + \epsilon \big) \odot \hat{\Delta}^2(\theta_h, i)$\footnote[1]{Consider that where $J(\theta_j, i) \approx 1$, $-\textbf{ln}\big(J(\theta_j, i)\big) \approx 0$, so where $J(\theta_j, i)$ is large, the difference $\hat{\Delta}^2(\theta_h, i)$ does not contribute significantly to loss. Alternatively, where$J(\theta_j, i) \approx 0$, $-\textbf{ln}\big(J(\theta_j, i)\big) \gg 0$, meaning that any substantial value in $\hat{\Delta}^2(\theta_h, i)$ will contribute heavily. A small additive $\epsilon = 0.001$ is used to prevent numerical errors.} 
\end{center}
\begin{flushleft}
The loss term contributed by masked squared ``pseudo-background" error loss as the element-wise mean of the masked squared differentials:
\end{flushleft}
\begin{center}
$\mathcal{L}_h(\theta_h, \theta_j, i) = \textbf{mean}\big(\hat{\mathcal{L}}_h(\theta_h, \theta_j, i)\big)$
\end{center}
\begin{flushleft}
Including the mean quantity masked with a multiplicative hyper-parameter:
\end{flushleft}
\begin{center}
$\mathcal{L}(\theta_h, \theta_j, i) = \mathcal{L}_h(\theta_h, \theta_j, i) + \alpha \cdot \textbf{mean}\big(J(\theta_j, i)\big)$\footnote[2]{This ``pixel-wise cost" addition to the entropic loss essentially forces the model to make a decision; increase loss via an expenditure associated with masking out more pixels, or increase loss by choosing not to mask out error between the background and the input at a particular location. Thus, the model will be incentivized to mask out error where and only where entropic loss would be expected to exceed $\alpha$, which, with an accurate background estimation, would only be at the location of a moving object.}
\end{center}
Which serves as the loss function for both Jekyll and Hyde. This unified loss function allows for the computer to execute a single back-propagation operation per training batch, which speeds up computation appreciably.

\section{HYPER-PARAMETERS USED AND TRAINING REGIME}
\label{sec:train}

Initially, the simulated data is sliced into samples with $N = 16,\ W = 64,\ H = 64$, and separated into Training, Validation, and Testing subsets with 50\%, 20\%, and 30\% of the samples in them respectively, with a total of approximately 1,000 samples over all three subsets. It was found experimentally that $\alpha = 1$ produced high quality results, where $\alpha$ is the multiplicative hyper-parameter associated with assigning pixel-wise cost to masking. \\
\\
Jekyll, Hyde, and Utterson were each given 100 epochs with PyTorch's native implementation of Adam optimizer \cite{kingma2014method, paszke2019pytorch}. All three models had a weight decay of 0.01, Hyde and Utterson had a learning rate of $5.0*10^{-4}$, and Jekyll had a learning rate of $5.0*10^{-5}$. All other hyper-parameters were left as default. Jekyll's learning rate was made lower than Hyde's as it was found that, learning at the same rate, Jekyll would begin masking out large errors that Hyde produced early in training, with Hyde never learning to correct the errors, and Jekyll never being able to stop masking them out. Jekyll and Hyde were trained using the loss function described earlier in this paper, while Utterson was trained with vanilla Binary Cross-Entropy Loss.

\section{RESULTS}
\label{sec:results}

\begin{figure}[b]
\caption{\textit{The Positive Predictive Values (PPV) and Negative Predictive Values (NPV) by threshold on the left and right, respectively. PPV is a measure of the likelihood that, if the model has labeled a pixel as containing a target, it does actually contain a target. NPV measures the likelihood that if the model says there is no target there, that it does not contain a target. The threshold (t) determines at what point the numerical output by the model is considered to be a positive prediction, e.g. for $t = 0.3$, if the model outputs a 0.4, then it is considered to be predicting the presence of a target, and if it outputs 0.2, it is read as predicting the absence of a target.}}

\includegraphics[height=0.3\textwidth]{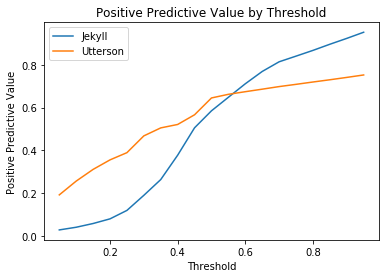}
\includegraphics[height=0.3\textwidth]{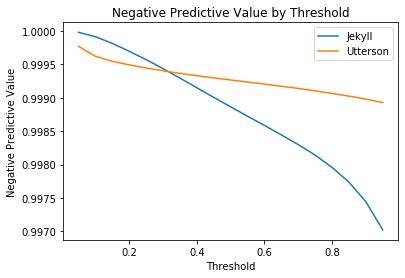}
\centering
\label{fig:ppvnpv}
\end{figure}

\begin{figure}[b]
\caption{\textit{The Sensitivity and Specificity by threshold on the left and right, respectively. Sensitivity measures, of the pixels that do actually contain a target object, the proportion of which that were actually picked up by the model. Specificity measures, of the pixels that do} not \textit{contain a target object, the proportion that was appropriately ignored by the model.}}
\includegraphics[height=0.3\textwidth]{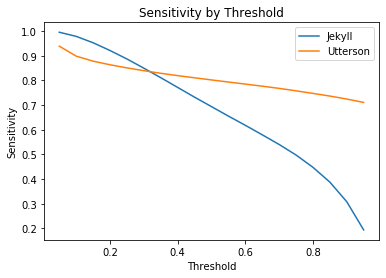}
\includegraphics[height=0.3\textwidth]{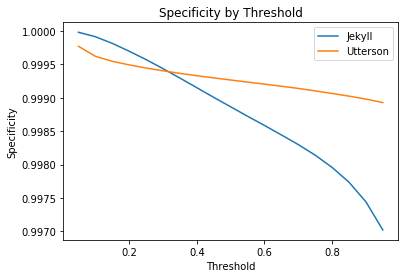}
\centering
\label{fig:sensspec}
\end{figure}

Numerical results describing the accuracy of the target tracking models, the unsupervised Jekyll and the supervised Utterson, on the testing data-set are reported in \textit{Figures \ref{fig:ppvnpv} and \ref{fig:sensspec}}.\footnote[1]{Due to the nature of the differences in approach to evaluation by the supervised and unsupervised models, Utterson was incapable of producing an output exceeding $\sim 0.45$. To account for this in numerical and graphical comparisons, Utterson's outputs were linearly re-scaled to $(0, 1)$, similar to those of Jekyll.} These were obtained for a variety of ``threshold" values, by, for a given threshold $t$, considering any model output greater than or equal to $t$ to be a positive prediction (i.e. predicting the presence of a target), and any output less than $t$ to be a negative prediction (i.e. predicting the absence of a target). As the threshold increases, a model is less likely to be read as predicting the presence of a target as the model must be more ``confident" for this to be the case, but it is also less likely to raise a false alarm. The numerical results include: Positive Predictive Value, the likelihood of a model's positive predictions to be correct; Negative Predictive Value, the likelihood of a model's negative predictions to be correct; Sensitivity, the likelihood for a model to pick up a particular target pixel; and Specificity, the likelihood for a model to reject a particular non-target pixel.\footnote[2]{These numerical results are meant as a rough comparison between models, and as support for the claims made. They do not precisely represent the quality or value of the outputs of the models, e.g. both models tend to over-represent the size of the target objects, as can be seen in the included imagery, lowering the PPV significantly, while still accurately representing the locations and motions of targets. Of these results, Sensitivity is the most meaningful, as it provides a measurement of the ability of the models to pick up targets.}\\

\begin{figure}[b]
\caption{\textit{Below is a figure showing each element of the Jekyll and Hyde formulation: an input series or ``video" of ``images" containing small moving objects, a static pseudo-background generated by one unsupervised model ``Hyde", a sequence containing the estimated objects detected by another unsupervised model ``Jekyll," Hyde's pseudo-background subtracted from the respective input frame, the results of a supervised model ``Utterson" for comparison, and the true labels/locations of the objects of interest. The first, second, third, and fourth rows relate to the $1^{st}, 5^{th}, 9^{th}$, and $13^{th}$ frames of a 16-frame input sequence respectively. In this instance, there are stationary features in this input sequence that closely resemble the target objects. The target tracking models demonstrate both their accuracy and value by ignoring these.}}
\includegraphics[width=0.8\textwidth]{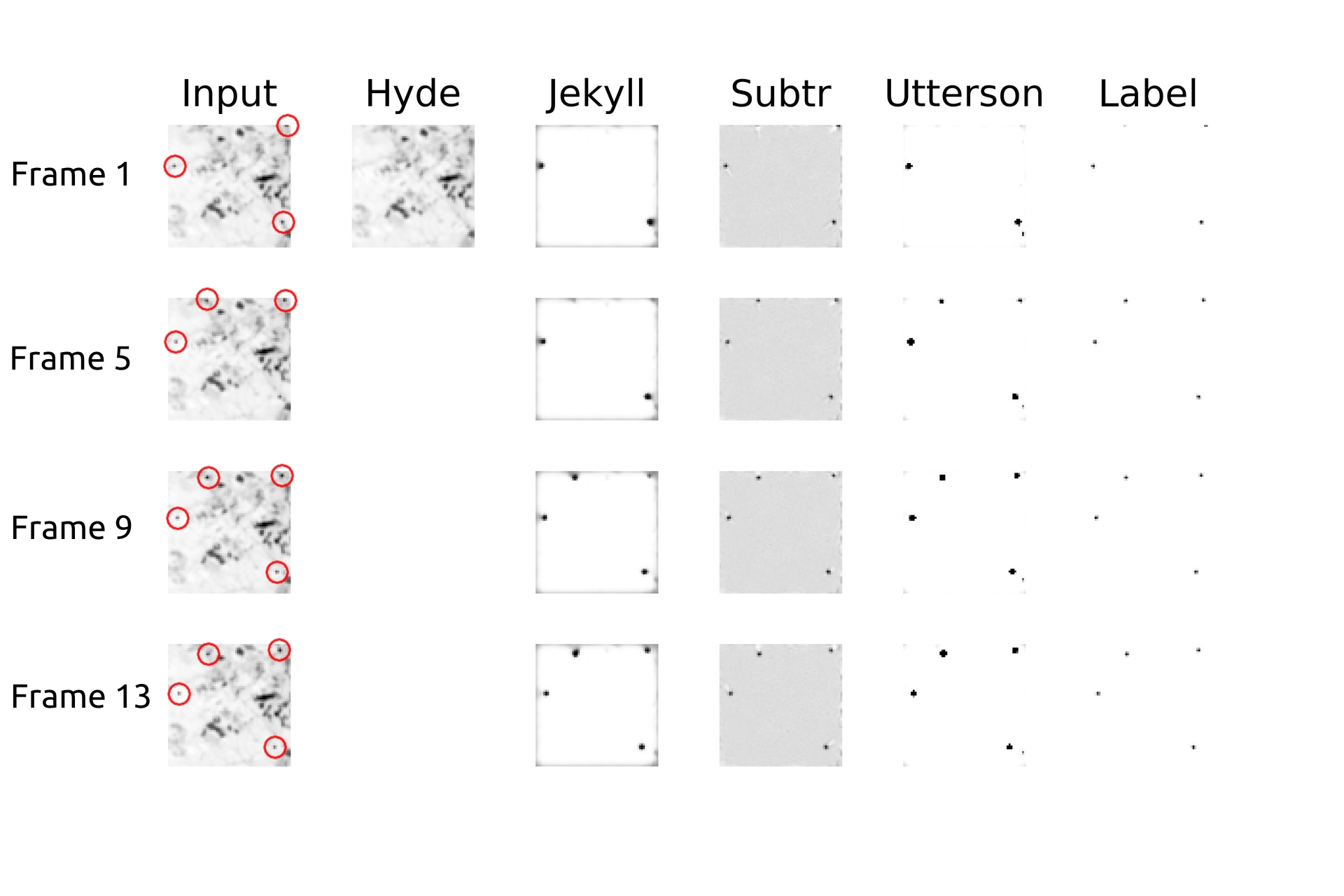}
\centering
\label{fig:res1}
\end{figure}

\begin{figure}
\caption{\textit{Here, the target tracking models manage to pick up a small, dark, very slowly moving object on a dark background near other dark objects. By looking closely at the locations marked out by the models, one can just barely make out the target, which would be otherwise unnoticed.}}
\includegraphics[width=0.8\textwidth]{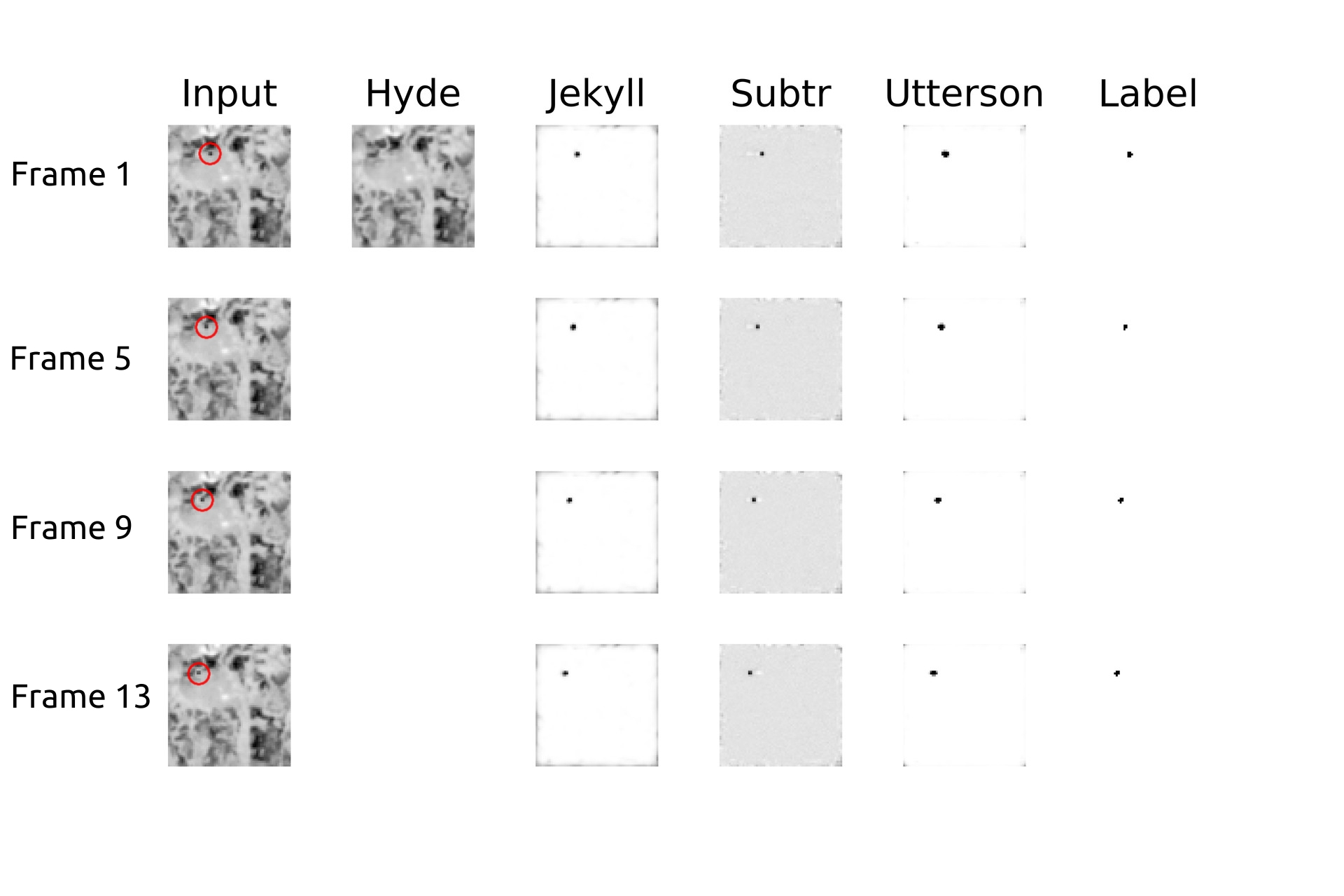}
\centering
\label{fig:res2}
\end{figure}

\begin{figure}
\caption{\textit{This example shows the models' ability to track a large number of objects at the same time, while also being able to ignore parts of the background that superficially resemble the objects being tracked.}}
\includegraphics[width=0.8\textwidth]{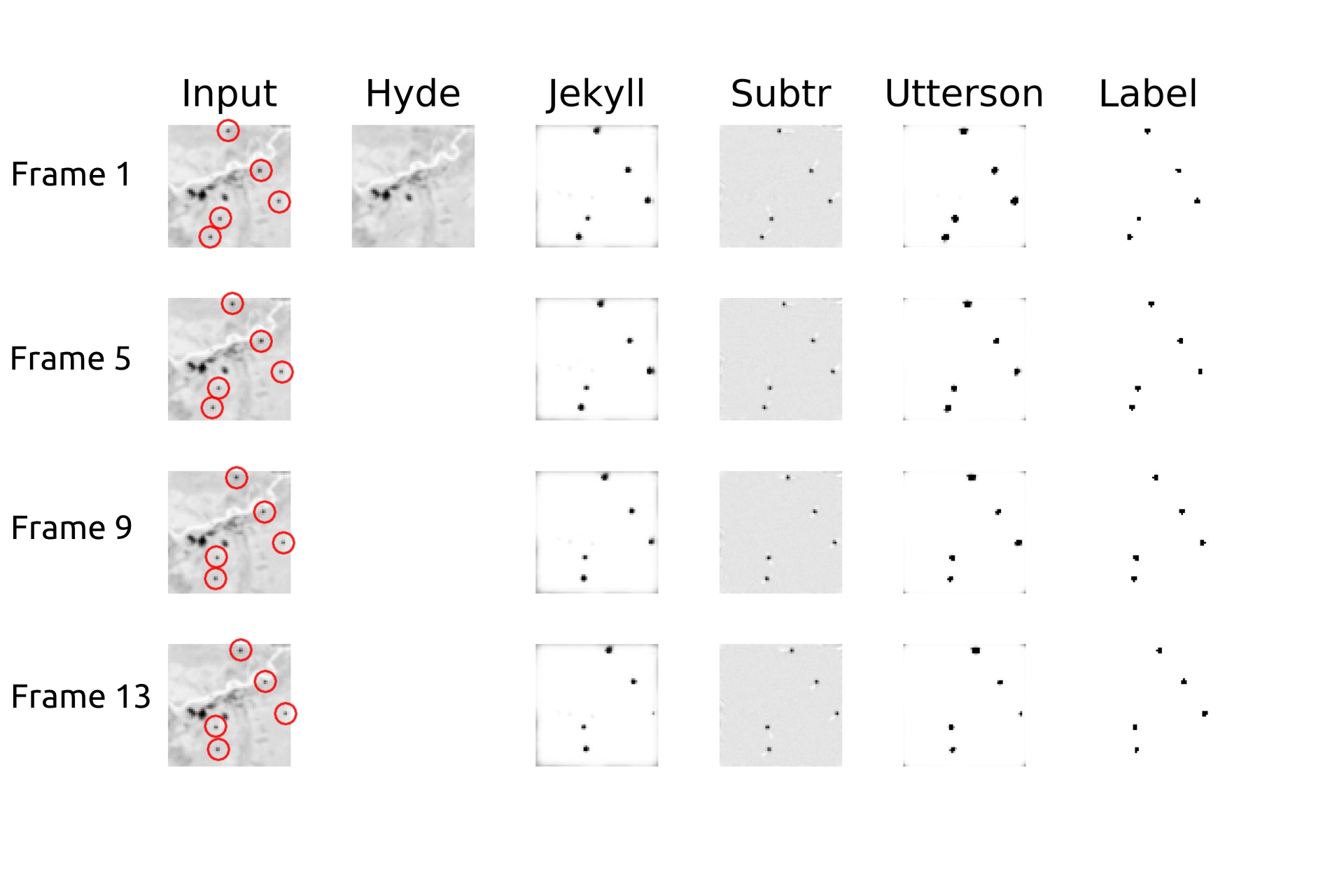}
\centering
\label{fig:res3}
\end{figure}

\begin{figure}
\caption{\textit{The models are capable not only of accurately tracking moving objects, but also of accurately producing a negative result.}}
\includegraphics[width=0.8\textwidth]{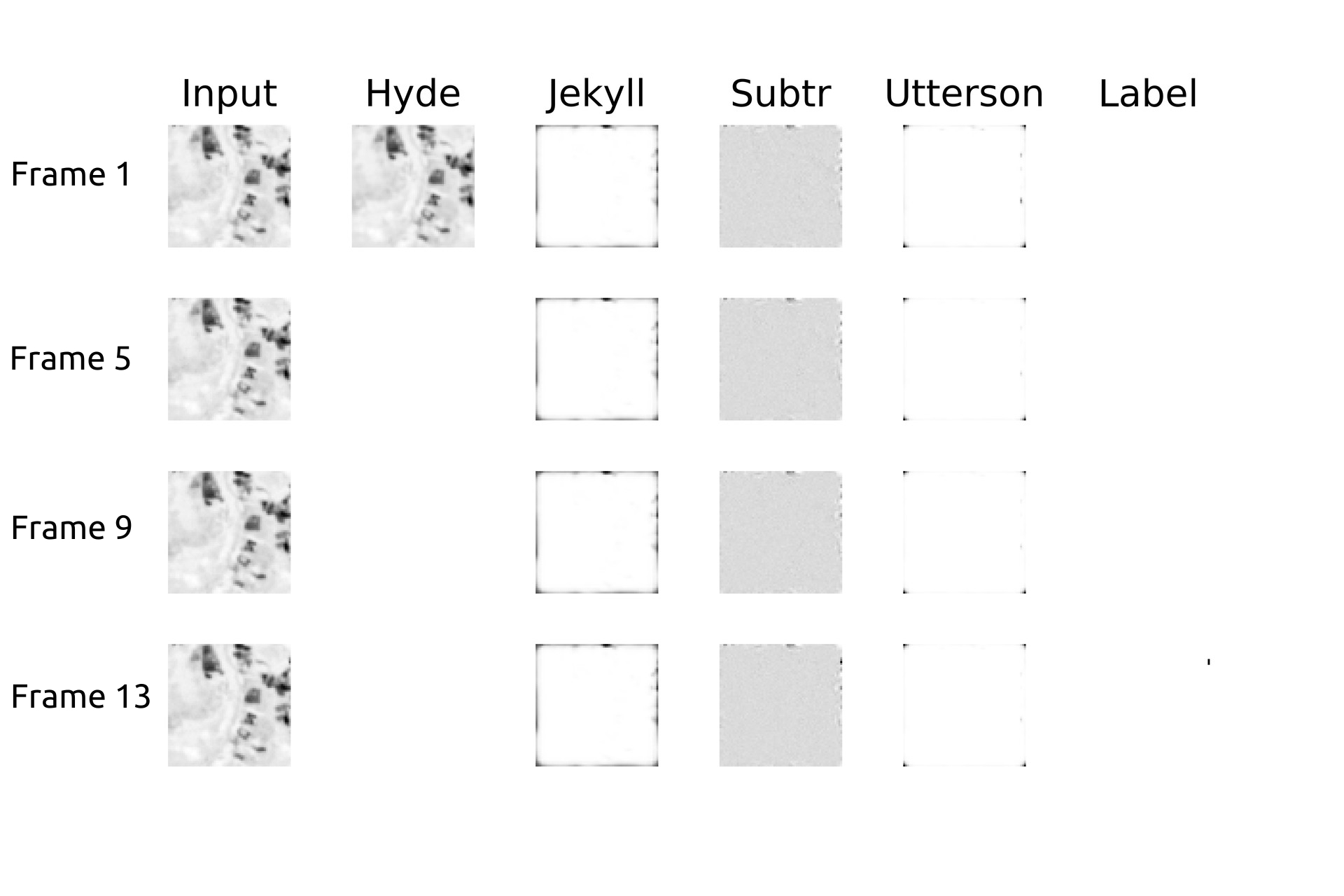}
\centering
\label{fig:res4}
\end{figure}

\begin{figure}
\caption{\textit{An example of the limitations of the models being used; when two tracked objects come close to one other, the model will track them both as a single misshapen object or ``blob", rather than as discrete entities. This is the result of both Jekyll and Utterson giving ``slack" to target objects, and masking them as larger than they are, as a precaution against costly failure to mask something.}}
\includegraphics[width=0.8\textwidth]{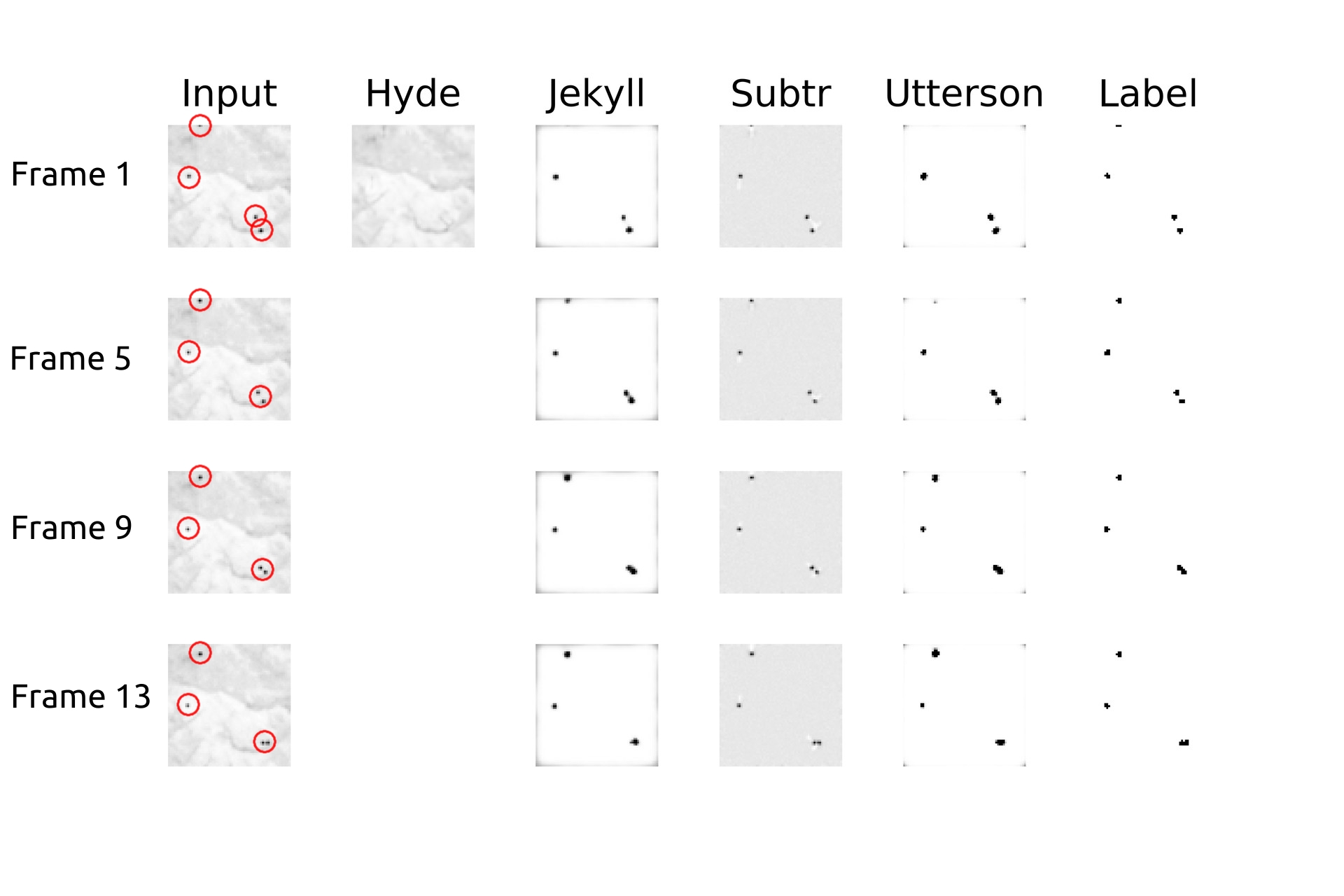}
\centering
\label{fig:res5}
\end{figure}

\noindent In addition, imagery is provided in \textit{Figures \ref{fig:res1}, \ref{fig:res2}, \ref{fig:res3}, \ref{fig:res4}, and \ref{fig:res5}} demonstrating the results. Of the column titles: ``Input" refers to the input passed to Jekyll, Hyde, and Utterson; ``Hyde", ``Jekyll", and ``Utterson" each show the respective outputs of those models; ``Subtr" shows the result of subtracting Hyde's pseudo-background from the input; and ``Label" shows the ground truths, the locations of targets of interest that were synthetically generated alongside the data. The samples shown come from the test subset of the data, and show frames 1, 5, 9, and 13 of a 16 frame sequence, with frame 1 of each sequence on the top of the figure, with the time increasing in each subsequent row.\\
\\
\noindent It can clearly be seen in the graphical results that the unsupervised approach demonstrated by Jekyll and Hyde is competitive with the supervised approach represented by Utterson, within this particular problem domain on this data-set. This is supported by the similarity of the Sensitivity values between Jekyll and Utterson for thresholds below $0.4$.

\section{CONCLUSIONS AND FUTURE WORK}
\label{sec:conc}

While this formulation of target tracking as an unsupervised learning problem is potentially extremely powerful, this particular approach is limited in scope. On its own, it is not robust to transient sensor artifacts, to background motion, or to weather. It also does not perform any sort of discrimination between targets worth tracking and those not worth tracking, classification of different kinds of targets, or prediction of trajectories.\\
\\
However, this work does present an important step towards the development of a completely unsupervised pipeline, by allowing for the transformation of imagery into target locations. Artifact removal \cite{winkler2011automatic, allman2017machine, biswas2013application, zhang2017unsupervised}, image stabilization \cite{walha2013video, saitwal2016image, deng2007airborne}, and saliency estimation \cite{li2010probabilistic, tang2016saliency, xia2016bottom, hu2018unsupervised, aytekin2018probabilistic} are all active research areas making great strides, and the coalescing of these under an unsupervised or semi-supervised Machine Learning paradigm would allow for an extremely robust, accurate, scalable, and inexpensive solution for a variety of challenges in satellite imagery analysis.

\begin{center}
\scalebox{.2}{\textcolor{white}{``A little song, a little dance, a little seltzer down your pants." -Tim}}
\end{center}

\bibliography{report} 
\bibliographystyle{spiebib} 

\end{document}